\documentclass{ecai}
\usepackage{graphicx}
\usepackage{latexsym}
\usepackage{amsmath}

\usepackage{tikz}
\newcommand{\tikzxmark}{%
\tikz[scale=0.23] {
    \draw[line width=0.7,line cap=round] (0,0) to [bend left=6] (1,1);
    \draw[line width=0.7,line cap=round] (0.2,0.95) to [bend right=3] (0.8,0.05);
}}

\def\checkmark{\tikz\fill[scale=0.4](0,.35) -- (.25,0) -- (0.8,.7) -- (.25,.15) -- cycle;} 

\begin{document}

\begin{frontmatter}
\title{
{\small \textcolor{red}{The full version of this paper appeared in proceedings of ECAI 2023}}

The Problem of Coherence in Natural Language Explanations of Recommendations}

\author[A]{\fnms{Jakub}~\snm{Raczyński}}
\author[A,B]{\fnms{Mateusz}~\snm{Lango}\thanks{Corresponding Author. Email: mlango@cs.put.poznan.pl.}\orcid{0000-0003-2881-5642}}
\author[A]{\fnms{Jerzy}~\snm{Stefanowski}\orcid{0000-0002-4949-8271}} 

\address[A]{ Poznan University of Technology,  Faculty of Computing and Telecommunications, Poznan, Poland}
\address[B]{Charles University, Faculty of Mathematics and Physics, Prague, Czech Republic}

\begin{abstract}

Providing natural language explanations for recommendations is particularly useful from the perspective of a non-expert user. 
Although several methods for providing such explanations have recently been proposed, we argue that an important aspect of explanation quality has been overlooked in their experimental evaluation.  
Specifically, the coherence between generated text and predicted rating, which is a necessary condition for an explanation to be useful, is not properly captured by currently used evaluation measures.  In this paper, we highlight the issue of explanation and prediction coherence by 1) presenting results from a manual verification of explanations generated by one of the state-of-the-art approaches 2) proposing a method of automatic coherence evaluation 3) introducing a new transformer-based method that aims to produce more coherent explanations than the state-of-the-art approaches 4) performing an experimental evaluation which demonstrates that this method significantly improves the explanation coherence without affecting the other aspects of recommendation performance.
\end{abstract}

\end{frontmatter}

\section{Introduction}

With the recent development of artificial intelligence comes a growing awareness of the risks involved and the need for greater control over such systems. In particular, the inability to explain the predictions of complex machine learning systems (usually treated as black boxes) is detrimental because it complicates debugging, hinders bias identification, prevents users from gaining trust in AI, etc.~\cite{samekmuller}.


The interest in \textit{explainable AI} also includes \textit{recommendation systems},  since providing good explanations for the recommendations can increase their effectiveness and improve the satisfaction of users by allowing them to better understand the system's predictions \cite{90344716}.  Among the various approaches to this task, generating \textit{explanations in natural language} is of particular interest, as such explanations are easier for the user to understand, require little or no prior user training, and can be potentially used to support a dialogue with the user. 

Modern approaches for this kind of explainable recommendation increasingly use deep neural models, as they produce richer and more fluent textual explanations than the earlier methods based on predefined sentence templates~\cite{90344716}. 
Many works~\cite{ACL21-PETER,CIKM20-NETE,TOIS23-PEPLER} evaluate new explainable recommendation methods in a rather comprehensive way, taking into account more than a dozen of performance aspects such as recommendation quality, text fluency, and personalization of the explanation provided to a given user. 
Nevertheless, these studies rely exclusively on automatic evaluation metrics such as BLEU~\cite{papineni-etal-2002-bleu} or ROUGE~\cite{lin2004rouge} and do not include evaluation by human users.

This motivated us to  carry out  a preliminary study\footnote{See the online appendix for details: \url{https://www.cs.put.poznan.pl/mlango/publications/ecai23.pdf}} in which we manually analysed a small random sample of explanations provided by one of the recently proposed state-of-the-art approaches, PETER+~\cite{ACL21-PETER}.
As to be expected, some flaws were observed in terms of text fluency such as repetition of the same words, unnatural sentence endings, or generations of sentences that lacked the context to be understood. However, the most frequently observed critical problem was the inconsistency between the recommender's prediction and the text of the generated explanation (see Fig.~\ref{fig:example2}). For instance, the text \textit{the characterizations are very good} surprisingly was sometimes provided as an explanation for the lowest rating. In fact, for some datasets such inconsistent explanations occurred even for 40\% of analyzed instances (see Sec.~\ref{sec:eval-coh}). Note that some of the aforementioned problems regarding text quality can be tackled by using more advanced natural language generation (NLG) methods (e.g. other decoding algorithms~\cite{su2022a}), but the critical issue of lack of coherence between textual explanation and prediction is inherent to the problem of prediction explainability and can not be solved with standard NLG methods. 

Therefore, despite the use of a wide range of performance measures in related studies, this critical factor concerning the quality of the generated explanations has not yet been sufficiently explored. 
The issue is further illustrated by the example presented in Fig.~\ref{fig:example}, where two explanations are provided for the same, low assessment of the alignment between a movie and user's preferences. 
The first textual explanation describes the film in glowing terms and actually suggests watching it, which is inconsistent with the predicted low rating and possibly causes user confusion.
The second explanation is coherent with the predicted rating and expresses a negative assessment of the movie.
It is worth noting that the otherwise perfect textual explanations which do not match the predicted outcome are not only entirely incorrect but also undermine the trust of the user in using an AI-based system.
Unfortunately, this issue has been overlooked in previous research, since the currently used evaluation measures separately assess predicted ratings and generated explanations. Therefore, they do not compare the predicted rating and generated explanation.

\begin{figure}
    \centering
    \begin{tabular}{|p{14mm}ll|}
    \hline
       Prediction (out of 5) & Explanation&Coh.\\\hline\hline
         5&this is a wonderful film & \checkmark\\
         4&   it's a goofy comedy that isn't funny  &\tikzxmark\\
1& the cast is very good  &\tikzxmark \\\hline
5 &they have a great drink selection  &\checkmark\\
5&the parking lot is always full&\tikzxmark\\
2&the staff is very friendly and helpful&\tikzxmark\\

   \hline
    \end{tabular}
    \caption{Predictions and their explanations generated by PETER+ for selected instances from Amazon Movies (up) and Yelp (down). For some, the lack of prediction-explanation coherence (Coh.) can be observed.}
    \label{fig:example2}
\end{figure}

\begin{figure}
    \centering
    \begin{tabular}{|ll|}
    \hline
         Predicted rating:& 2 (out of 5)\\
PETER+ explanation (SOTA): & it 's a fun movie  \\
   CER explanation (this work): & it is a waste of time \\
   \hline
    \end{tabular}
    \caption{Explanations generated by PETER+ and CER for an instance from Amazon Movies. Both methods predicted the same rating in this case.}
    \label{fig:example}
\end{figure}

Guided by the above observations, we focus our attention on the problem of \textit{coherence} between generated text and the predicted recommendation output, which is a necessary condition for an explanation to be useful. 
Addressing this issue properly requires, on the one hand, the development of an automatic method for evaluating the prediction-explanation consistency, which will enable simple evaluation of current and future explainable recommendation methods without the time-consuming manual data annotation. 
On the other hand, a new recommendation system generating personalized and consistent explanations in natural language should be proposed. 
In particular, the main contributions of our paper are as follows:
\begin{enumerate}
    \item carrying out a manual evaluation of reference explanations from the popular datasets as well as explanations generated by state-of-the-art methods, drawing research attention towards the problem of coherence between natural language explanations  and predicted ratings in the recommendation domain,
\item introducing a new trainable, reference-less metric for automatic coherence evaluation of the predictions and explanations, 
\item proposing a new transformer-based method that aims to generate more coherent explanations through a new intermediary task of explanation-based recommendation,
\item performing experiments with three benchmark datasets, where we compare our method against other state-of-the-art methods and  show that it improves the explanation coherence without decreasing other measures of the recommendation predictions.

\end{enumerate}

\section{Related work}


Following \cite{90344716}, the methods for explaining the prediction of recommender systems can be categorized into six groups based on the type of algorithm being used, i.e. association rules, factorization models, topic modeling, knowledge graphs, agnostic post-hoc explanation methods and finally deep neural networks, which are at the focus of this work.
Many of the deep learning solutions for explaining predictions use various types of attention mechanism~\cite{vaswani2017attention}.  For example, the combination of convolutional layers and attention mechanism presented in~\cite{Seo} allowed the identification of phrases in user opinions that had the greatest influence on the recommendation. Attention mechanisms are also used in deep recommendation systems that operate on multi-modal data, e.g. marking important areas of the product image as an explanation~\cite{chen2019personalized}. 
Other versions of networks based on encoder-decoder architecture using GRU or LSTM modules are used to analyze logs of subject-user interactions~\cite{tao2019log2intent}. 

Finally, the methods that use neural networks to generate explanations in natural language also belong to this group.
Such approaches include post-hoc methods as attribute-to-sequence model (Att2Seq)~\cite{dong-etal-2017-learning-generate} that can be used as a separate module explaining recommendations. The method uses LSTM networks to generate a product review basing on the expected rating and representations of a user and a product.
Another similar method is ACMLM~\cite{ni2019justifying} which uses a dedicated aspect decoder to guide explanation generation performed by a fine-tuned BERT language model~\cite{devlin-etal-2019-bert}.
The idea of exploiting pre-trained large language models was also applied in a recent PEPLER approach~\cite{TOIS23-PEPLER} that takes advantage of prompt-based transfer learning with GPT-2 model~\cite{radford2019language}.

Other methods are specifically designed to perform the recommendation task along with providing textual explanations.
One such method is NRT~\cite{li2017neural} which jointly predicts ratings and generates so-called tips using a recurrent neural network having only the embedding of a user and item as the input.
Yet another proposal is NETE~\cite{CIKM20-NETE}, which additionally uses information about the item's feature to personalize explanations. 
The method generates text with GRU-inspired recurrent units, the inner workings of  which can be interpreted as the generation of a neural sentence template.


PETER+~\cite{ACL21-PETER} is another recent neural recommender that generates personalized natural language explanations, obtaining state-of-the-art results.
In contrast to the previously mentioned NRT and NETE methods, it is based on the transformer architecture~\cite{vaswani2017attention} that jointly produces deep feature representations for both rating prediction and text generation.
The model is general enough to be trained and to provide explanations without item features (PETER), but achieves much better personalization of the generated explanations while using them (PETER+). Given the promising experimental results, it was selected for further consideration in this paper.
It is worth noting that recently, \cite{xie2022faithfulness} also pointed out some issues regarding the factuality and \emph{semantic} coherence of generated explanations by several methods, including PETER (see the discussion of semantic and prediction-explanation coherence 
in the appendix).

\section{Measuring coherence between explanation and prediction}
\label{sec:meas-coh}


\paragraph{Problem statement}
 Given a set of users $U$, a set of items $I$ and lists of item features $F_{u,i}$ that can be used to justify a recommendation, the task of explainable recommendation is understood in this paper as the task of jointly predicting a rating $r_{u,i}$ and an explanation $E_{u,i}$.
Let's further assume that the rating  is a score $r_{u,i} \in \{1,2,3,4,5\}$ measuring the alignment of an item with the user's preferences, while the explanation $E_{u,i}=[ e_1,e_2,...,e_n ]$ is a sequence of tokens from the vocabulary $V$ justifying the score assigned by the recommender.
The generated textual explanation should not only be personalized for a given user by mentioning or referring to the selected item features\footnote{Note, that the set of item features $F_{u,i}$ depends not only on the item $i$ but also on the user $u$, since in this formulation $F_{u,i}$ contains only the item features which are relevant (important) for the user $u$.}  $F_{u,i}$ but it should also be semantically consistent with the predicted rating $r_{u,i}$.



Recall that in the related works~\cite{ACL21-PETER,CIKM20-NETE,TOIS23-PEPLER,SIGIR17-NRT}, the quality of  explanations $E$ is assessed with different measures, that can be  divided into three categories: the measures that verify the personalization of generated explanations by computing various statistics of item feature's mentions; standard metrics for assessing the quality of predicted ratings and metrics comparing $n$-grams from generated texts and reference explanations. However, none of these metrics jointly analyses the predicted rating and generated explanation, therefore overlooking the problem of prediction-explanation coherence.

Measuring the degree of coherence between the predicted rating and the generated textual explanation is challenging for various reasons.
First, the measure should combine a numerical input (rating) with a textual one (explanation).
Next, it cannot straightforwardly use the gold standard data as a reference. Although the explanation contained in the training data is supposedly\footnote{Our manual analysis revealed that actually,  25-30\% of explanations in the commonly used datasets are not coherent with the rating, indicating that the issue of recommendation-explanation coherence was not properly handled even in the data collection process -- see Sec. \ref{sec:eval-coh} for details.} coherent, it is only coherent with the gold standard rating, so a disparity between the predicted ratings and the gold standard ratings may invalidate the coherence of the gold standard explanation. Moreover, standard metrics like BLEU~\cite{papineni-etal-2002-bleu}, which compare generated text with the reference through $n$-grams, do not distinguish between replacing a word in the reference text to its synonym (preserving coherence) and changing it to a random word.

In our study, we first assess the coherence between generated explanations and predicted ratings by manually annotating a random sample of predictions obtained from models. Later, using the annotated data, we put forward the proposal of a trainable, reference-less coherence metric that allows automatic evaluation of coherence for large datasets. Below we describe these approaches while the experimental results are presented in Section \ref{sec:exp}.

\paragraph{Manual verification}
The process of manual annotation of the obtained predictions was performed by two independent human annotators to which we presented pairs of predicted ratings and explanations $(r_{u,i}, E_{u,i})$ and asked to binary assess its coherence (coherent/ not coherent). 
The binary assessment of the level of coherence is a certain simplification, but it is recommended to perform evaluations of models' interpretability with relatively simple tasks performed by humans~\cite{lage2019human}.
Additionally, the binary assessment of coherence allowed for simpler annotation rules, minimizing the inconsistencies that may occur in manual evaluation~\cite{zeng2021validating}, enabled faster expert training and accelerated the annotation process.

While evaluating model predictions, the predicted scores were rounded to integers to keep the data format from the datasets.
For ratings equal to 1 and 5, the annotators were instructed to treat as coherent only the explanations mentioning solely positive or negative characteristics of an item.
For ratings 2 and 4, mentioning a minor disadvantage/advantage of an item being evaluated was also considered consistent with the rating. Specifically, minor disadvantages and advantages were identified by phrases such as "slightly", or "little bit". 
For a rating of 3, neutral sentences, very mildly polarized with positive or negative item features, mentioning both the good and bad sides of the item, as well as explanations from which it was not possible to deduce what opinion their potential author could have about the subject were treated as coherent.


Following these rules both annotators provided coherence labels for 100 randomly selected examples from datasets used in the experiments (see Section~\ref{sec:exp}).
The inter-annotator agreement was of 92\%. 
After the following discussion of label inconsistency and clarification of annotation rules, the inter-annotator agreement increased to 96\%.
Due to the relatively high inter-annotator agreement, in the remaining experiments with the manual evaluation of models' performance, each example was annotated only once to reduce the annotator's workload. The results of the final manual evaluation of explanations and predictions using 1800 instances from various models are discussed in Section~\ref{sec:eval-coh}.



\paragraph{Trainable coherence metric}

Performing manual evaluation is a tedious and time-consuming task, making it impractical to evaluate more than a limited sample of recommender predictions. 
Nevertheless, constructing an automatic metric assessing the prediction-explanation coherence is challenging since the evaluation seems to require a general understanding of the generated text, handling synonyms, etc.
Similar evaluation problems have been observed in dialogue agents and machine translation communities, which have addressed them by adopting various trainable metrics~\cite{tao2018ruber,blanc}.
Such metrics employ a machine learning model that learns to evaluate generated texts, e.g. by mimicking the decisions of human evaluators. Such models can be trained on a limited number of provided human judgments for some models and datasets, but once trained they can be used to evaluate different models on different datasets.

Inspired by these works, we put forward a proposition of an automatic, trainable metric for coherence evaluation.
The proposed metric consists of a binary classifier trained on our annotated examples to assess the coherence between the generated explanation and the  predicted rating.
The classifier is later applied to all the model's predictions and the percentage of predictions marked by the classifier as coherent is treated as a performance indicator.

The proposal of a coherence classifier should address two problems: the limited size of training data and the heterogeneous input consisting of a number (rating) and a text (explanation).
We address both these problems by exploiting the potential of pretrained large language models (LLM) and converting the whole problem into a sentence classification task.

For each of the possible ratings, we developed a sentence template which after being filled with the text of explanation constitutes an input to the classifier. The task of the classifier is to  simply detect the correctness of such formed sentence.
This setup of the problem enables effective utilization of knowledge acquired by an LLM during pretraining.
The sentence templates for each rating are as follows:
\begin{enumerate}
    \item An example of very negative review is 
\item An example of slightly negative review is
\item An example of neutral or mixed review is
\item An example of slightly positive review is
\item An example of very positive review is
\end{enumerate}
For instance, to verify the consistency of PETER+ prediction from the example depicted in Fig.~\ref{fig:example}, the input to the LLM-based classifier should be "An example of slightly negative review is it 's a fun movie". 
The training of the coherence classifier is performed by optimizing the binary cross-entropy loss.

\section{Generating more coherent explanations}
\label{CER}

\begin{figure}
\centerline{  \includegraphics[width=\linewidth]{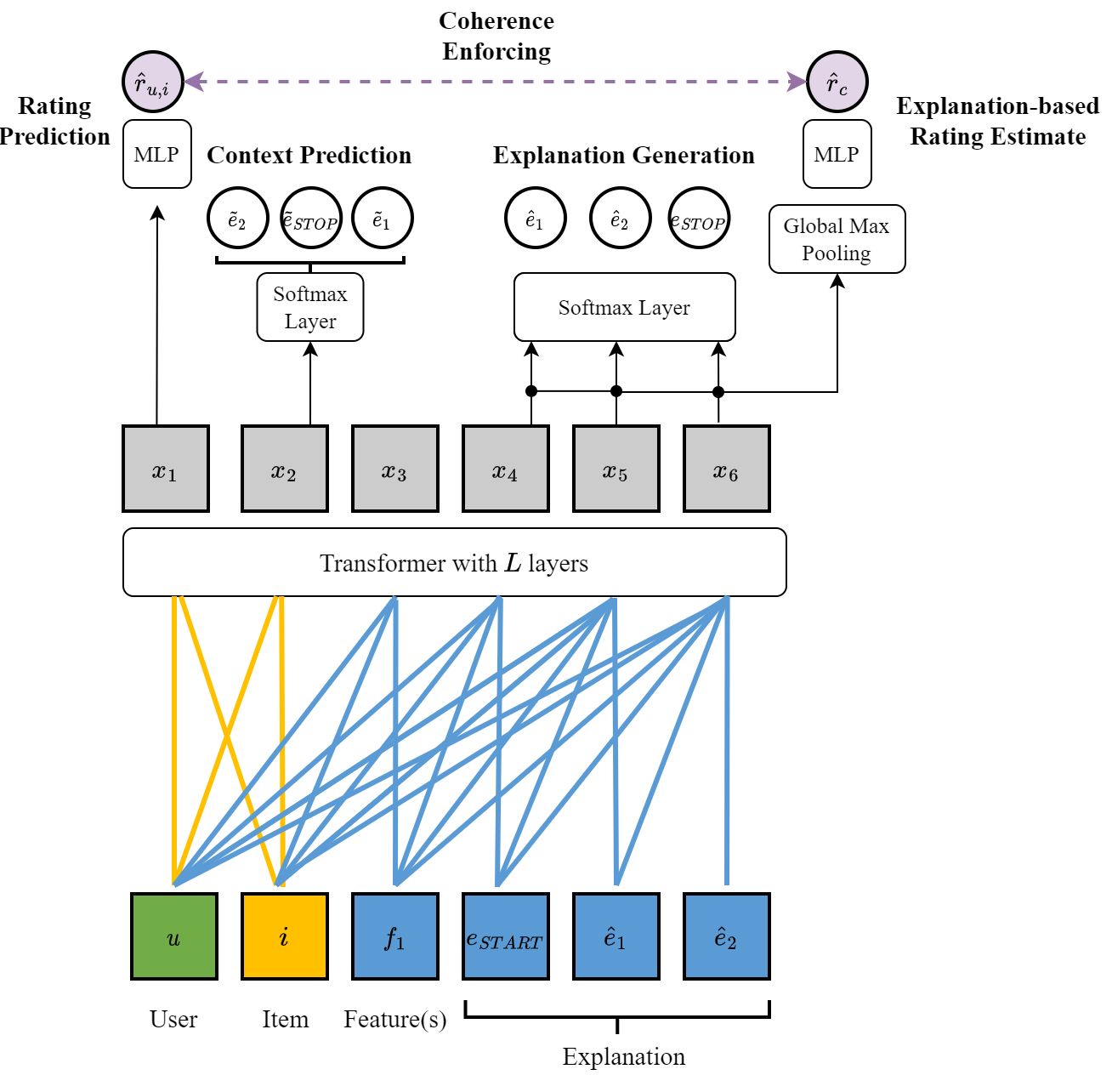}}
  \caption{An overview of the proposed Coherent Explainable Recommender (CER) architecture.}
  \label{fig:peter}
\end{figure}

We introduce a new method for explainable recommendation called Coherent Explainable Recommender (CER) that builds upon the state-of-the-art PETER+~\cite{ACL21-PETER} architecture and extends it with additional mechanisms ensuring better coherence between the predicted rating and the generated explanation.
This is achieved by adding to the architecture a special neural module and putting forward a new associated intermediary task of explanation-based rating estimation.
The overview of the proposed architecture is depicted in Figure~\ref{fig:peter}.

\paragraph{Architecture of the proposed model} 


The backbone of the proposed CER method is a transformer module~\cite{vaswani2017attention}, which effectiveness for the natural language generation tasks has been confirmed in many works~\cite{radford2019language,gpt3}.
The input to the transformer layers is designed to allow the unified construction of the deep representations needed for all subsequent prediction tasks. 
More precisely, a sequence 
$$S=[u,i,f_1, f_2, ..., f_{|F_{u,i}|}, e_{START},e_1, e_2, ..., e_n]$$
containing representations of the user $u$, item $i$, item features $f_j\in F_{u,i}$ and the explanation $e_j \in E_{u,i}$ is formed.
Both users and items are represented by distributed feature vectors stored in embedding matrices $U$ and $I$, respectively. 
Similarly, the representations of each token from the vocabulary $V$, that are used to express the item features as well as words of the recommendation explanation, are stored in an embedding matrix $E$.
The embedding of the special token $e_{START}$, which marks the beginning of explanation generation, is also stored in this matrix.

The input sequence is passed to a transformer layer with self-attention mechanism defined as 
$$X = softmax\left(\frac{ SW_qW_k^TS}{\sqrt{d}} +M  \right)SW_v$$
where $W_q,W_k,W_v$ are weight matrices, softmax function is performed row-wise and $M$ is the masking matrix~\cite{vaswani2017attention} which controls which elements of the input sequence can attend to which elements. 
Following~\cite{ACL21-PETER}, we use a masking matrix that generally allows attending all the previous elements of the sequence and forbids attending all elements to the right of the considered input element, with one exception. 
While constructing the output representation for the user (the first element of the sequence) the neural network is allowed to attend to the item embedding (i.e., the second sequence element). 
This enables the construction of the output embedding, which contains information about both the user and the item, and can be used as the basis for performing the recommendation.

Our model contains multiple stacked transformer layers, with the number of layers being a model hyperparameter.
The aim of using several transformer layers is to construct a deep contextual representation of each input token, which serves as input to four training tasks performed by fully-connected networks.
These tasks are: \emph{rating prediction}, \emph{explanation generation}, \emph{context prediction}, and \emph{explanation-based rating estimation}.
The first two of these tasks, rating prediction and explanation generation, are target tasks, i.e. they are performed during both training and testing phases and provide the expected output of an explainable recommender. The remaining two tasks are intermediary tasks that provide an additional training signal. 
Context prediction task is to improve the personalization of the generated explanations, whereas the new explanation-based rating estimation task aims to provide additional training signal encouraging the generation of coherent explanations. They are described in  the following paragraphs.
Despite the fact that our model has several transformer layers, in the following descriptions we will slightly abuse the notation and refer to the output of the last transformer layer as $X$.

\newcommand{\Lagr}{\mathcal{L}}

\paragraph{Rating prediction} 
Since the focus of this work is not to improve the quality of the provided recommendations, the proposed model uses a rather simple two-layered fully-connected network to predict the rating $\hat{r}_{u,i}$. 
The input to this network is the output representation of the user $x_1$, but note that due to the modification of the masking mechanism, the representation of $x_1$ is computed while also considering the item $i$.
The used loss function is the mean-squared error (MSE) defined as follows:
$$\Lagr_r = \frac{1}{|D|} \sum_{(u,i) \in D} \left( \hat{r}_{u,i} - r_{u,i} \right)^2 $$
$$\hat{r}_{u,i} = w_r^T\sigma (W_rx_1 +b_r)$$
where $D$ is the training dataset, $w_r, W_r, b_r$ are learnable parameters and $\sigma()$ denotes sigmoid activation function.

\paragraph{Explanation generation} 
The explanation is generated using a greedy algorithm, that selects the words to which a softmax layer assigned  the highest probability. The training objective is  the cross-entropy function.
$$\Lagr_e = - \frac{1}{|D|} \sum_{(u,i) \in D} \frac{1}{|E_{u,i}|} \sum_{j=1}^{|E_{u,i}|}  \log \text{softmax}(W_ex_{|F_{u,i}|+j+2}+b_e)_{e_j}$$
where $D$ is the training dataset, $W_e,b_e$ are weights of the softmax layer and $e_j$ is the $j$-th token of gold standard explanation $E_{u,i}$.
Note, that the index of $|F_{u,i}|+j+2$ simply selects the representation of $e_j$ from the output sequence $X$.

\paragraph{Context prediction}
The auxiliary task of context prediction is to estimate the probability distribution of all words occurring in the explanation based solely on the representation $x_2$, i.e. a representation constructed while attending user and item representations only. 
This task promotes better entanglement between the predicted explanation and the representations of users and items, resulting in more personalized explanations. 
In the proposed model, the output probability distribution is computed by a softmax layer and trained through cross-entropy optimization.
$$\Lagr_c = - \frac{1}{|D|} \sum_{(u,i) \in D} \frac{1}{|E_{u,i}|} \sum_{j=1}^{|E_{u,i}|}  \log \text{softmax}(W_cx_2+b_c)_{e_j}$$
where $D$ is the training dataset, $W_c,b_c$ are learnable weights and $e_j$ is the $j$-th token of gold standard explanation $E_{u,i}$.

\paragraph{Explanation-based rating estimation}
The goal of this intermediary task is to promote better coherence between the generated explanation and the predicted rating.
We claim that basing solely on a coherent explanation expressed in natural language, it should be possible to guess the predicted rating with a reasonable accuracy.
For instance, an explanation that points out weaknesses of the item and uses negatively polarized words should be a strong indicator that the recommendation head should have predicted a low rating. 
Similarly, an explanation focused on the positive aspects of an item should be associated with a positive rating. 
Therefore, we propose using an additional recommendation head that predicts rating $r_{u,i}$ while having only the representation of the generated explanation as its input.

To obtain a fixed-sized representation of textual explanation, we apply a max pooling over time~\cite{kim-2014-convolutional} operator on the word embeddings of the generated text. 
$$ \widetilde{E}_{u,i} = \text{Max-pooling}\left(x_{|F_{u,i}|+3}, x_{|F_{u,i}|+4},...,x_{|F_{u,i}|+|E_{(u,i)}|+2}\right)$$
where Max-pooling() returns a vector filled with the maximum values computed over each dimension of the input vectors. 
Later, the auxiliary rating score is predicted with a two-layered MLP network with a linear output.
$$\hat{r}_{E_{u,i}} = w_{coh}^T\sigma (W_{coh} \widetilde{E}_{u,i}  +b_{coh})$$
where $w_{coh}, W_{coh}, b_{coh}$ are learnable weights of MLP and $\sigma()$ denotes the sigmoid activation function.
The optimized loss function enforces the alignment between the rating predicted by the recommendation head $\hat{r}_{u,i}$ and the rating estimated solely from the text of explanation $\hat{r}_{E_{u,i}}$.
For this purpose, the mean squared error defined below is used as the loss function.
$$\Lagr_{coh} = \frac{1}{|D|} \sum_{(u,i) \in D} \left( \hat{r}_{u,i} - \hat{r}_{E_{u,i}} \right)^2 $$
where $D$ is the training dataset, $\hat{r}_{u,i}$ is the predicted rating by the recommendation head and $\hat{r}_{E_{u,i}}$ is the predicted rating basing only on the generated explanation.

Note that solving this intermediary task provides an additional training signal to the model that incentivizes the generation of coherent explanations. The auxiliary prediction of rating is ignoring information about the user $u$ and item $i$ and is exclusively based on the generated text. Thus, the model must reflect the rating in the text and generate a coherent explanation in order to succeed in this task. 
Moreover, the text-based rating prediction head is trained to mimic the predictions of the recommendation head and not to reflect the gold standard rating, which further increases coherence by promoting explanations aligned with the actual prediction and not with the gold standard.

\paragraph{Joint loss function} Finally, the whole CER model is trained with the standard backpropagation algorithm that optimizes a joint loss function defined as a sum of all four prediction tasks.
$$\Lagr = \Lagr_{r}+\Lagr_{e}+\Lagr_{c}+\Lagr_{coh}$$
where $\Lagr_{r},\Lagr_{e},\Lagr_{c},\Lagr_{coh}$ are loss functions of particular tasks as defined in previous paragraphs.

\section{Experimental evaluation}
\label{sec:exp}

\subsection{Experimental setup}

To verify the utility of the proposed Coherent Explainable Recommender approach, we conducted experiments on three datasets provided by~\cite{CIKM20-NETE} which are typically used to evaluate explainable recommenders: Amazon Movies, TripAdvisor, and Yelp. 
The original train-test splits provided by the dataset's authors were also used.
For a fair comparison, we mimicked the experimental setup of the work that originally introduced PETER+ method~\cite{ACL21-PETER}. 
This included comparison to the same baselines, employing the same quality metrics (except new ones proposed in this paper) and using the same hyperparameters for the CER approach (like the number of transformer layers, embedding size, number of neurons in hidden layers etc.).
The only difference is within the methods compared, as we added recently released, GPT-2 based PEPLER~\cite{TOIS23-PEPLER} to the experiments. The code is available on our GitHub repository\footnote{\url{https://github.com/JMRaczynski/CER}}

The quality of explanation generated by the proposed CER method was compared to other approaches which take item features into account such as PETER+~\cite{ACL21-PETER}, NETE~\cite{CIKM20-NETE} and ACMLM~\cite{ni2019justifying}, as well as explainable recommenders that do not make use of features: PEPLER~\cite{TOIS23-PEPLER}, Att2Seq~\cite{dong-etal-2017-learning-generate}, NRT~\cite{SIGIR17-NRT} and standard Transformer~\cite{vaswani2017attention} network trained by using identifiers of users and items as additional words. 
The recommendation performance was additionally compared to classical non-explainable recommenders: a probabilistic matrix factorization (PMF)~\cite{mnih2007probabilistic} and SVD++\cite{koren2008factorization}.

 A diversified collection of metrics is employed to assess the explainability and text quality of the generated explanations as well as the recommendation performance.
 The latter is measured with mean absolute error (MAE) and root mean squared error (RMSE) measures.
 The text quality is evaluated by comparing the generated explanations against references provided in the datasets using metrics like BLEU~\cite{papineni-etal-2002-bleu} and ROUGE~\cite{lin2004rouge}.
 More specifically, BLEU-1 (B1), BLEU-4 (B4) as well as precision (P), recall (R), and F-score (F) of ROUGE-1 (R1) and ROUGE-2 (R2) are reported.
 The explainability properties and personalization of the generated texts are assessed with metrics related to the usage of item features in the provided explanations.
 We employ Feature Matching Ratio (FMR), Feature Coverage Ratio (FCR), and Feature Diversity (DIV) metrics proposed in~\cite{CIKM20-NETE}.
  In addition, the Unique Sentence Ratio (USR), which measures the diversity of generated texts, is also reported.
  
   \begin{table*}[t]\tabsize
  \begin{center}
\caption{The comparison of the quality of the explanations generated by CER and by other explainable methods under study.}
\label{basicevaluation}
\begin{tabular}{ll|llllllllllll}
\hline
\multicolumn{1}{c}{}       &      & \multicolumn{3}{c|}{\textbf{Explainability}}   & \multicolumn{8}{c}{\textbf{Text Quality}}  \\ 
    &  & FMR & FCR & \multicolumn{1}{l|}{DIV$\downarrow$} & USR & B1 & B4 & R1-P & R1-R & R1-F & R2-P & R2-R & R2-F \\ \hline
    
\multicolumn{1}{l}{} &  & \multicolumn{12}{c}{\textbf{Yelp}}  \\ \hline
\multicolumn{2}{l|}{Transformer} & 0.06 &0.06&\multicolumn{1}{l|}{2.46} & 0.01 & 7.39 & 0.42 & {19.18} & 10.29 & 12.56 & 1.71 & 0.92 & 1.09 \\
\multicolumn{2}{l|}{NRT} & 0.07 &0.11&\multicolumn{1}{l|}{2.37} & 0.12 & {11.66} & 0.65 & 17.69 & 12.11 & 13.55 & 1.76 & 1.22  & 1.33 \\
\multicolumn{2}{l|}{Att2Seq} & 0.07 &0.12&\multicolumn{1}{l|}{2.41} & 0.13 & 10.29 & 0.58 & 18.73 & 11.28 & 13.29 & 1.85 & 1.14  & 1.31 \\
\multicolumn{2}{l|}{PEPLER} & {0.08} &{0.30}&\multicolumn{1}{l|}{{1.52}} & {0.35} & 11.23 & {0.73} & 17.51 & {12.55} & 13.53 & 1.86 & {1.42} & 1.46 \\ 

\multicolumn{2}{l|}{ACMLM} & 0.05 & 0.31 &\multicolumn{1}{l|}{0.95} & \underline{0.95} & 7.01 & 0.24 & 7.89 & 7.54 & 6.82 & 0.44 & 0.48 & 0.39 \\
\multicolumn{2}{l|}{NETE} & 0.80 &0.27&\multicolumn{1}{l|}{1.48} & 0.52 & 19.31 & 2.69 & 33.98 & 22.51 & 25.56 & 8.93 & 5.54 & 6.33 \\
\multicolumn{2}{l|}{PETER} & {0.08} & 0.19 &\multicolumn{1}{l|}{1.54} & 0.13 & 10.77 & {0.73} & 18.54 & 12.20 & {13.77} & {2.02} & 1.38 & {1.49} \\

\multicolumn{2}{l|}{{PETER+}} & \underline{0.87} &0.31&  \multicolumn{1}{l|}{\underline{0.94}} & 0.20 & \underline{20.71} & \underline{3.75} & 34.17 & \underline{26.45} & 27.64 & 10.12 & \underline{7.92} & \underline{7.97} \\
\multicolumn{2}{l|}{CER} & 0.86 & \underline{0.37} &\multicolumn{1}{l|}{1.08} & 0.30 & 20.62 & 3.42 & \underline{35.51} & 26.03 & \underline{27.92} & \underline{10.74} & 7.43  & \underline{7.97} \\
\hline \hline

  & & \multicolumn{12}{c}{\textbf{Amazon Movies}}  \\ \hline
\multicolumn{2}{l|}{Transformer} & 0.10 & 0.01 &\multicolumn{1}{l|}{3.26} & 0.00 & 9.71 & 0.59 & 19.68 & 11.94 & 14.11 & 2.10 & 1.39  & 1.55 \\
\multicolumn{2}{l|}{NRT} & {0.12} & 0.07 &\multicolumn{1}{l|}{2.93} & 0.17 & 12.93 & 0.96 & {21.03} & 13.57 & {15.56} & 2.71 & 1.84  & 2.05 \\
\multicolumn{2}{l|}{Att2Seq} & {0.12} &0.20&\multicolumn{1}{l|}{2.74} & 0.33 & 12.56 & 0.95 & 20.79 & 13.31 & 15.35 & 2.62 & 1.78 & 1.99 \\

\multicolumn{2}{l|}{PEPLER} & 0.11 & {0.27} &\multicolumn{1}{l|}{2.06} & {0.38} & {13.19} & 1.05 & 18.51 & {14.16} & 14.87 & 2.36 & 1.88 & 1.91 \\ 

\multicolumn{2}{l|}{ACMLM} & 0.10 & \underline{0.31} &\multicolumn{1}{l|}{2.07} & \underline{0.96} & 9.52 & 0.22 & 11.65 & 10.39 & 9.69 & 0.71 & 0.81 & 0.64 \\
\multicolumn{2}{l|}{NETE} & 0.71 & 0.19 &\multicolumn{1}{l|}{1.93} & 0.57 & 18.76 & 2.47 & 33.87 & 21.43 & 24.81 & 7.58 & 4.77 & 5.46 \\
\multicolumn{2}{l|}{PETER} & {0.12} &0.21&\multicolumn{1}{l|}{{1.75}} & 0.29 & 12.77 & {1.17} & 19.81 & 13.80 & 15.23 & {2.80} & {2.08} & {2.20} \\
\multicolumn{2}{l|}{{PETER+}} & \underline{0.80} &0.23&  \multicolumn{1}{l|}{\underline{1.14}} &0.25& 17.20 & \underline{3.13} & \underline{35.43} & 23.40 & 26.22 & \underline{9.32} & 6.13  & 6.75 \\ 
\multicolumn{2}{l|}{CER} & 0.78 & \underline{0.31}&\multicolumn{1}{l|}{1.24} & 0.44 & \underline{19.88} & 3.12 & 34.98 & \underline{24.22} & \underline{26.60} & 9.24 & \underline{6.37} & \underline{6.86} \\
\hline \hline

  &  & \multicolumn{12}{c}{\textbf{TripAdvisor}} \\ \hline
\multicolumn{2}{l|}{Transformer} & 0.04 &0.00&\multicolumn{1}{l|}{10.00} & 0.00 & 12.79 & 0.71 & 16.52 & {16.38} & 15.88 & 2.22 & {2.63} & {2.34} \\
\multicolumn{2}{l|}{NRT} & 0.06 &0.09&\multicolumn{1}{l|}{4.27} & 0.08 & 15.05 & 0.99 & 18.22 & 14.39 & 15.40 & 2.29 & 1.98 & 2.01 \\
\multicolumn{2}{l|}{Att2Seq} & 0.06 &0.15&\multicolumn{1}{l|}{4.32} & 0.17 & 15.27 & 1.03 & 18.97 & 14.72 & 15.92 & 2.40 & 2.03 & 2.09 \\

\multicolumn{2}{l|}{PEPLER} & {0.07} & {0.21} &\multicolumn{1}{l|}{{2.71}} & {0.24} & 15.49 & 1.09 & 19.48 & 15.67 & 16.24 & {2.48} & 2.21 & 2.16 \\ 

\multicolumn{2}{l|}{ACMLM} & 0.07 & \underline{0.41} &\multicolumn{1}{l|}{\underline{0.78}} & \underline{0.94} & 3.45 & 0.02 & 4.86 & 3.82 & 3.72 & 0.18 & 0.20 & 0.16 \\
\multicolumn{2}{l|}{NETE} & 0.78 &0.27&\multicolumn{1}{l|}{2.22} & 0.57 & 22.39 & 3.66 & 35.68 & 24.86 & 27.71 & 10.20 & 6.98 & 7.66 \\
\multicolumn{2}{l|}{PETER} & {0.07} &0.13&\multicolumn{1}{l|}{2.95} & 0.08 & {15.96} & {1.11} & {19.07} & 16.09 & {16.48} & 2.33 & 2.17 & 2.09 \\
\multicolumn{2}{l|}{{PETER+}} & \underline{0.91} &0.38&  \multicolumn{1}{l|}{1.51} & 0.26 & \underline{25.68} & \underline{5.02} & 34.97 & \underline{30.36} & 30.26 & 10.75 & \underline{9.48} & 9.13 \\
\multicolumn{2}{l|}{CER} & 0.88 &0.39&\multicolumn{1}{l|}{1.62} & 0.32 & 24.66 & 4.61 & \underline{37.04} & 29.34 & \underline{30.42} & \underline{11.84} & 9.02 & \underline{9.24} \\\hline
\end{tabular}
\end{center}
\end{table*}
\subsection{Evaluation of explanation quality}

Although our work is focused on improving the coherence of the explanations provided by the recommender, we want to make sure that the proposed modifications do not negatively influence other aspects of the generated explanations. So, we first evaluated explanations generated by CER with typical text quality and explainability metrics used in the related works. The results of these experiments and the comparison with other related methods can be found in Table~\ref{basicevaluation}.

The obtained scores demonstrate that our model generates explanations that are at least on par with those provided by the original PETER+ architecture. 
In terms of both ROUGE-1 F-score and ROUGE-2 F-score, which aggregate precision and recall, CER obtains the highest scores for all datasets under study.
While for some datasets BLEU scores obtained by CER are slightly lower than those obtained by PETER+, the differences are not large and the USR measure shows the superiority of the proposed approach for all datasets.
These results could indicate that CER, by addressing the issue of coherence, also generates texts of slightly better quality than PETER+.

Regarding the measures of explainability, there is a trade-off between a higher feature coverage provided by CER (FCR) and better feature precision and diversity (FMR, DIV) provided by PETER+. 
In conclusion, as observed differences are small, presented results show comparable explanation quality of CER and PETER+.

\subsection{Evaluation of coherence between explanation and predicted rating}
\label{sec:eval-coh}

The coherence between the predicted rating and the generated explanation was evaluated using the methods introduced in Section~\ref{sec:meas-coh}.
In the experimental evaluation, we compare the coherence obtained by the proposed CER model and its predecessor PETER+.
In addition, we also evaluated the ground truth prediction-explanation pairs, as the datasets were originally constructed with explanations heuristically extracted from user reviews by a method described in~\cite{CIKM20-NETE}. 
The heuristic dataset construction process implies the possible occurrence of noisy examples and thus may contain some incoherent examples. 

\paragraph{Manual evaluation}
We first proceeded with the manual evaluation of generated explanation and predicted rating pairs by two independent human annotators.
From each dataset we randomly selected 200 examples and performed the prediction by both models. 
In this way we obtained 600 rating-explanation pairs\footnote{including reference explanations from the dataset} for manual annotation for each dataset (1800 in total).
To avoid some potential biases, exactly half of rating-explanation pairs for each dataset and each model were annotated by one annotator and the other half by the other. The results of this evaluation are shown in Table \ref{manualresults}. 

\begin{table}
\begin{center}
{\caption{The results of manual analysis of coherence between explanations and predicted ratings for selected methods.}
\label{manualresults}}
\begin{tabular}{l|ccc}
\hline
                  & \multicolumn{3}{c}{\% of coherent explanations}              \\
                  & Yelp & {Amazon} & TripAdvisor \\     \hline
{PETER+}  & {65,0}  & {60,0}  & 82,5  \\
{CER} & \underline{68,0} & \underline{63,5} & \underline{89,0} \\
{Gold standard}    & {73,5}   & {70,0}  & 74,0    
    \\\hline
\end{tabular}
\end{center}
\end{table}

\begin{table*}[h]
\tabsize
  \begin{center}
{\caption{The results of automatic coherence evaluation of explanations and predicted ratings for selected methods.}
\label{aggregatedautomaticeval}}
\begin{tabular}{c|c|cccccccccc|c}
\hline
\textbf{Dataset} & {\textbf{Evaluated architecture}} & \multicolumn{10}{c|}{\textbf{Evaluating model}}                              &                 \\ \cline{3-13} 
                              &                                              & 1     & 2     & 3     & 4     & 5     & 6     & 7     & 8     & 9     & 10    & Mean  \\     \hline
{Yelp}       & PETER+    & 81.16 & 87.53 & 77.73 & 88.41 & 86.4 & 88.43 & 81.77 & 89.05 & 91.02 & 50.55 &  \textbf{82.21}      \\
                              & CER  & \underline{82.64} & \underline{88.58} & \underline{79.69} & \underline{89.33} & \underline{87.49} & \underline{89.42} & \underline{83.33} & \underline{89.93} & \underline{91.82} & \underline{53.22} &  \textbf{\underline{83.55}} \\
                              & Gold standard                                          & 90.17 & 94.72 & 88.72 & 95.28 & 93.83 & 94.93 & 90.8 & 95.96 & 96.5 & 73.89 & \textbf{91.48} \\ \hline 
{Amazon}      & PETER+ & 86.24 & 55.81 & 88.94 & 47.65 & 37.33 & \underline{78.49} & 73.87 & \underline{62.97} & 63.24 & 83.13 &   \textbf{67.77} \\
                              & CER & \underline{86.39} & \underline{56.21} & \underline{89.29} & \underline{48.27} & \underline{38.06} & 78.18 & \underline{74.12} & 62.95 & \underline{63.62} & \underline{83.25} & \textbf{\underline{68.03}} \\
                              & Gold standard    & 89.31 & 57.82 & 92.03 & 48.56 & 36.18 & 79.78 & 77.71 & 65.3 & 67.82 & 86.13 &  \textbf{70.07}  \\ \hline 
{TripAdvisor}      & PETER+    & \underline{92.84}     & \underline{93.99}     & \underline{90.02} & \underline{84.33}    & \underline{95.28}     & \underline{91.1}     & \underline{91.37}    & \underline{92.6}    & \underline{79.13}    & \underline{94.81}     & \textbf{\underline{90.55}}     \\
                              & CER  & 91.49 & 92.78 & 88.63 & 83.26 & 94.41 & 89.43 & 89.88 & 91.41 & 77.56 & 93.84  & \textbf{89.27} \\
                              &Gold standard  & 87.71 & 89.77 & 85.35 & 78.88 & 91.76  & 84.13 & 81.52 & 88.06 & 69.1 & 91.72 & \textbf{84.78} \\    \hline
\end{tabular}
\end{center}
\end{table*}

Surprisingly, the manual analysis  revealed a significant fraction of incoherent explanation-rating pairs in the datasets typically used to evaluate explainable recommenders.
The percentage of references coherent with the gold standard prediction ranged from 70\% in the Amazon Movies dataset to 74\% in the TripAdvisor one. 
Considering the quality of used datasets, both PETER and CER models did a good job in generating coherent results, producing on TripAdvisor dataset even more coherent explanation-rating pairs than the gold standard ones.
In terms of coherence, CER outperformed PETER+ on each dataset. The score differences in favor of CER are up to 6,5 percentage points on TripAdvisor dataset and 3 percentage points on the largest Yelp dataset.
The relative reduction of PETER+ inconsistency ranges from 8 to 37 percent.

\paragraph{Automatic evaluation}
With the encouraging results of the manual evaluation, we continued with the automatic coherence evaluation.
We applied the methodology of trainable coherence metric introduced in Section~\ref{sec:meas-coh} and constructed a dataset for sentence classification from our annotated data.
We fed these to the transformer-based pre-trained language model BERT~\cite{devlin-etal-2019-bert} and, following the transfer learning methodology advised by the model authors, we built a classification head on top of the representation of CLS token that marks the beginning of the sentence in this model.
The classification head consists of a fully-connected neural network with two layers. 
The network has 32 units in the hidden layer with hyperbolic tangent activation function. 
Since the constructed dataset is class imbalanced, the cross-entropy loss with class weights is optimized during training.
For each of the datasets, we tuned classifier hyperparameters in a 10-fold cross-validation process\footnote{The chosen values of hyperparameters are presented in the online appendix}.
To obtain more stable results, the reported values of the automatic coherence metric are averaged over 10 models trained with identical architecture.

The averaged results of the automatic evaluation as well as results obtained for each separate classifier run are presented in Table~\ref{aggregatedautomaticeval}. 
Analyzing the scores obtained for Yelp dataset, we can clearly see the difference in coherence between predicted ratings and generated explanations by PETER+ and CER. 
Each of the ten evaluation models demonstrated that CER generates more coherent explanations than PETER+ architecture. 
Consistently with human evaluation and our expectations, there is a gap between the coherence of CER and the references from the dataset.

The difference in coherence measure on Amazon dataset is smaller than on the previously discussed Yelp dataset, but still, CER outperforms PETER+ according to 8 out of 10 evaluation models, while one of the other two models shows a negligible difference of only 0.02\%. 
Even though the absolute difference between averaged results for both models is not very large, it is worth noting that the relative improvement of CER over PETER+ stands for over 11\% of the difference between PETER+ and the gold standard.

Finally, on TripAdvisor dataset PETER+ seems to achieve slightly better results than CER. However, counterintuitively, the ground truth data is even less coherent than the output of both PETER+ and CER according to the automatic evaluation.  
Therefore, CER actually obtains results closer to the gold standard ones than PETER+.
This phenomenon could be partially explained by PETER/CER architectures rarely predicting low ratings (and thus rarely generating negative explanations) due to a high imbalance of this dataset in terms of ratings. 
This problem, combined with conflicting results of the manual and automatic evaluation, makes it difficult to draw clear conclusions regarding CER coherence on this particular dataset. 

\subsection{Evaluating recommendation performance}



The comparison of recommendation performance in terms of MAE and RMSE measures is presented in Table~\ref{receval}. 
The obtained results show that CER provides virtually identical recommendation quality as PETER+ on all datasets and  is also comparable to this offered by other baselines.  Therefore, the CER improvement of explanation coherence does not degrade the recommendation performance.

\begin{table}
\begin{center}
{\caption{The comparison of the recommendation performance of  CER, PETER+ and other baseline methods.}
\label{receval}}
\begin{tabular}{cc|cc|cc|cc}
   \hline
\multicolumn{2}{c|}{}         & \multicolumn{2}{c|}{Yelp} & \multicolumn{2}{c|}{Amazon} & \multicolumn{2}{c}{TripAdvisor} \\ 
\multicolumn{2}{c|}{}                          & RMSE        & MAE         & RMSE         & MAE          & RMSE           & MAE            \\ \hline
\multicolumn{2}{l|}{PMF}                       & 1.09        & 0.88        & 1.03         & 0.81         & 0.87           & 0.70           \\
\multicolumn{2}{l|}{SVD++}                     & \underline{1.01}  & \underline{0.78}  & 0.96         & 0.72         & 0.80           & 0.61           \\
\multicolumn{2}{l|}{NRT}                       & \underline{1.01}  & \underline{0.78}  & \underline{0.95}   & \underline{0.70}   & \underline{0.79}     & 0.61           \\
\multicolumn{2}{l|}{NETE}                      & \underline{1.01}  & 0.79        & 0.96         & 0.73         & \underline{0.79}     & \underline{0.60}     \\
\multicolumn{2}{l|}{{PETER+}}           & \underline{1.01}  & 0.79        & \underline{0.95}   & 0.71         & 0.81           & 0.62           \\
\multicolumn{2}{l|}{{{CER}}} & \underline{1.01}  & 0.79        & \underline{0.95}   & 0.72         & 0.81           & 0.63           \\     \hline
\end{tabular}
\end{center}
\end{table}

\section{Summary}

In this paper, we draw research attention to the problem of coherence between generated textual explanations and predicted ratings in the domain of recommendation systems.
We argue that despite the fact that a lack of coherence completely invalidates the provided explanation, this aspect of the methods has not been properly captured in the standard measures used so far in the experimental evaluation.
Moreover, it has been especially surprising that the conducted manual verification of explanation-prediction coherence revealed that as many as 30\% of the reference explanations present in the commonly used datasets are incoherent.
Such a level of noise regarding a critical aspect of explanation quality was quite unexpected to be found in the datasets used to construct theoretically more trustworthy recommendation systems.
We believe that our experiments highlight the need for the construction of new, less noisy benchmarking datasets for explainable recommendations.

Nevertheless, the problem of insufficient coherence of predictions made by currently proposed models is not entirely the result of training on noisy datasets, but also of a lack of proper handling of this issue in the designed systems.
In this paper, we propose an additional intermediary task of explanation-based rating estimation, which provides an additional training signal that drives the recommendation models toward producing explanations that are more consistent with the predicted rating.
The incorporation of this task to modern explainable recommenders led us to the proposal of Coherent Explainable Recommender architecture that obtains superior results in terms of the coherence between generated explanations and predicted ratings, at the same time not influencing negatively the quality of provided explanations and other measures of text quality.
Still, the explanations provided by CER may lack factuality, semantic coherence, or causality -- see the discussion of limitations in the appendix.


\clearpage
\ack J. Raczyński was supported by  0311/SBAD/0743. M. Lango was supported by the European Research Council (Grant agreement No. 101039303 NG-NLG). J. Stefanowski was supported by National Science Centre, Poland (Grant No. 2022/47/D/ST6/01770)
For the purpose of Open Access, the author has applied a CC-BY public copyright licence to any Author Accepted Manuscript (AAM) version arising from this submission.

\bibliography{ecai}
\end{document}